\documentclass[10pt,twocolumn,letterpaper]{article}

\usepackage{cvpr}              

\usepackage[accsupp]{axessibility}  

\usepackage{amsmath,amssymb,amsfonts}
\usepackage{algorithmic}
\usepackage{graphicx}
\usepackage{textcomp}

\usepackage{booktabs}
\usepackage{afterpage}
\usepackage{tabularx, colortbl}
\usepackage[includehead]{}
\usepackage{fancyhdr}
\usepackage{float}
\usepackage{tikz}

\usepackage{tabu}
\usepackage{siunitx}
\usepackage{caption}
\usepackage{multirow}

\usepackage{makecell}

\definecolor{samgray}{gray}{0.95} 

\usepackage{etoolbox}
\AtBeginEnvironment{table}{\small}
\AtBeginEnvironment{table*}{\small}

\newcolumntype{Y}{>{\centering\arraybackslash}X}

\definecolor{darkgreen}{RGB}{0, 128, 0}
\newcommand{\best}[1]{\textbf{\textcolor{darkgreen}{#1}}}
\newcommand{\second}[1]{\textcolor{blue}{#1}}

\definecolor{cvprblue}{rgb}{0.21,0.49,0.74}
\usepackage[pagebackref,breaklinks,colorlinks,allcolors=cvprblue]{hyperref}

\def\paperID{4} 
\def\confName{CViW}
\def\confYear{2026}

\title{BED-SAM2: Boundary-Enhanced-Depth SAM2 via Monocular Geometric Priors}

\author{Tyler Rust\\
University of Delaware\\
{\tt\small trust@udel.edu}
\and
Dara McNally\\
University of South Florida\\
{\tt\small daramcnally@usf.edu}
\and
Kyle O'Donnell\\
University of South Florida\\
{\tt\small kpod@usf.edu}
\and
Colin Kelly\\
DEVCOM Army Research Laboratory\\
{\tt\small colin.d.kelly6.civ@army.mil}
\and
Chandra Kambhamettu\\
University of South Florida\\
{\tt\small ckambhamettu@usf.edu}
}

\begin{document}
\maketitle
\begin{abstract}

Building upon the SAM2 vision foundation model for downstream segmentation, this study introduces Boundary Enhanced Depth (BED)-SAM2. The SAM2 Hiera encoder architecture is modified to directly encode monocular depth information from RGB images, thereby providing geometric cues that enhance object boundary delineation and facilitate the extraction of camouflaged object shapes. BED-SAM2 demonstrates competitive state-of-the-art performance across multiple salient and camouflaged object detection tasks with as few as five training epochs. The code can be found here: \url{https://github.com/TylerRust-1/BED-SAM2}.
\end{abstract}

\section{Introduction}

\begin{figure*}[ht]
    \centering
    \includegraphics[width=1\textwidth]{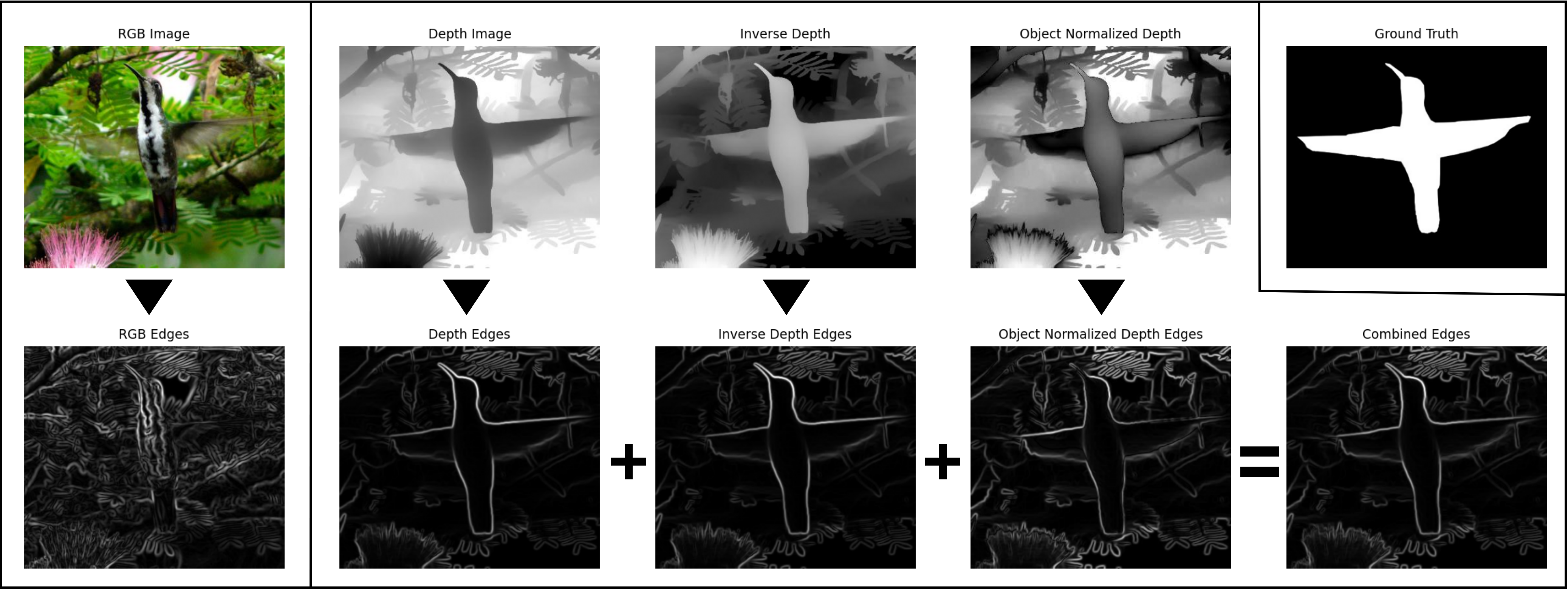}
    \caption{Cumulative structure map from monocular depth. Sobel filters are applied independently to the RGB, raw depth, inverse depth, and centered depth representations to extract edge maps. The final structure map is obtained by element-wise summation of all depth edge maps and is provided as a fourth channel input to the SAM2 Hiera backbone. The combined depth structure map provides clear boundary signals to the network unlike noisy RGB edges. Ground-truth and RGB segmentation is shown for reference.}
    \label{fig:bird_depth}
\end{figure*}

Since the early 2010s, models pre-trained on large-scale vision datasets such as ImageNet~\cite{Imagenet} and COCO~\cite{lin2015microsoftcococommonobjects} have become foundational in computer vision. Widely adopted architectures trained on these datasets include various ResNets~\cite{he2015deepresiduallearningimage} and VGG backbones~\cite{simonyan2015deepconvolutionalnetworkslargescale}. More recently, vision foundation models (VFMs) such as CLIP~\cite{radford2021learningtransferablevisualmodels} and SAM~\cite{kirillov2023segment} employ data-intensive Vision Transformers (ViTs)~\cite{dosovitskiy2021imageworth16x16words} to achieve strong zero-shot generalization.

While general-purpose segmentation models such as SAM and SAM2~\cite{kirillov2023segment,ravi2024sam2segmentimages} perform well on standard datasets, they struggle with fine-grained, low-level tasks such as camouflaged object detection (COD). To maintain model efficiency, recent research has focused on developing hybrid networks that combine the representational power of VFMs with the parallelization and fine-tuning efficiency of convolutional neural networks (CNNs) for downstream tasks with limited data~\cite{xiong2026sam2_unet}.

Salient object detection (SOD) is a binary image segmentation task that generates masks for visually prominent objects in images and videos. A central challenge in segmentation tasks is managing the precision-recall trade-off, specifically balancing comprehensive detection with the risk of over-detection. This challenge is particularly pronounced at decision boundaries, where variations in color and texture can reduce classifier confidence. Additionally, when an object is partially obscured by a non-salient foreground element, such as leaves in front of a bird, models often fail to accurately classify all edge pixels. This difficulty is further intensified in COD, where target objects are deliberately concealed, and decision boundaries are inherently ambiguous.

To overcome these limitations, we introduce Boundary Enhanced Depth-SAM2 (BED-SAM2), a hybrid VFM-based architecture for SOD that incorporates monocular depth estimation (MDE) to extract geometric cues from RGB images. Depth information delineates clear boundaries between foreground and background, reducing the impact of confounding factors such as texture and shadows. Through a novel early-fusion strategy, cumulative depth edges (as illustrated in Fig.~\ref{fig:bird_depth}) are integrated into the initial layer of the Hiera encoder, enabling state-of-the-art performance within as few as five training epochs. Our main contributions are as follows:

\begin{itemize}
    \item BED-SAM2, an efficient framework that integrates monocular depth priors into the SAM2 encoder to improve boundary adherence in fine-grained segmentation.
    \item We introduce a cumulative structure map strategy that aggregates multiple depth-derived features to deliver robust geometric cues and mitigate confounding effects from texture and color.
    \item We demonstrate that our early-fusion approach achieves state-of-the-art performance on $13$ benchmarks while significantly reducing training overhead, requiring as few as five epochs.
\end{itemize}

\section{Related Works}
SOD has evolved significantly from early heuristic methods that relied on handcrafted features. Deep CNN backbones such as VGG~\cite{simonyan2015deepconvolutionalnetworkslargescale} and ResNet~\cite{he2015deepresiduallearningimage} served as feature extractors for early dense prediction models. U-Net~\cite{ronneberger2015unetconvolutionalnetworksbiomedical_unet} introduced an efficient single encoder-decoder with skip-connections to capture multi-scale context features without relying on pre-trained backbones. BiRefNet~\cite{Zheng_2024_BiRefNet} uses the Swin backbone and employs a coarse-to-fine pipeline for segmentation by zooming in on complex patches to capture fine details. The proposed method, BED-SAM2, improves segmentation decision boundaries by incorporating distilled depth priors at the encoder stage. Modern approaches highlight the increasing emphasis on high-resolution fidelity, efficient computation, and the use of external priors to improve saliency predictions.

\subsection{Vision Foundation Models}
Meta AI's SAM model~\cite{kirillov2023segment} is trained on eleven million images with over one billion image masks. The network provides a single-pass image encoder and a lightweight mask decoder. SAM2 introduced the Hiera Backbone~\cite{ryali2023hierahierarchicalvisiontransformer}, a hierarchical vision transformer trained with a masked autoencoder (MAE).

SAM2-Large contains over 200 million parameters. To keep model fine-tuning lightweight, recent work, such as SAM2-Adapter~\cite{chen2024sam2adapterevaluatingadapting} inserts multi-layer perceptrons (MLPs) into the Hiera encoder. While the original image encoder is frozen, the adapters can be trained. These adapters achieve improved results on downstream tasks, with SAM2-Adapter achieving state-of-the-art (SOTA) performance across several challenging benchmarks (e.g., CAMO, CHAMELEON, COD10K). This performance comes at the cost of adding slight architectural complexity and training overhead to the encoder. Instead of the SAM2-Adapter, we opt for Low-rank adaptation (LoRA) fine-tuning, a technique that uses trainable rank-decomposition matrices to reduce complexity and save memory. Since dense segmentation requires modifying both global semantic attention and local spatial representations, LoRA preserves the pre-trained attention manifold with minimal latency. Adapter-based methods operate as residual bottlenecks and do not directly reshape attention maps.

\subsection{Monocular Depth Estimation (MDE)}
Depth estimation has traditionally relied on stereo vision techniques that utilize disparity between synchronized image pairs to triangulate depth~\cite{scharstein2002taxonomy,hirschmuller2007stereo}. Classical stereo methods, including block matching, semi-global matching, and graph-cut optimization, establish correspondences but encounter difficulties in textureless regions and occlusions, and require precise camera calibration.

\begin{figure*}[t!]
    \centering
    \includegraphics[width=\textwidth]{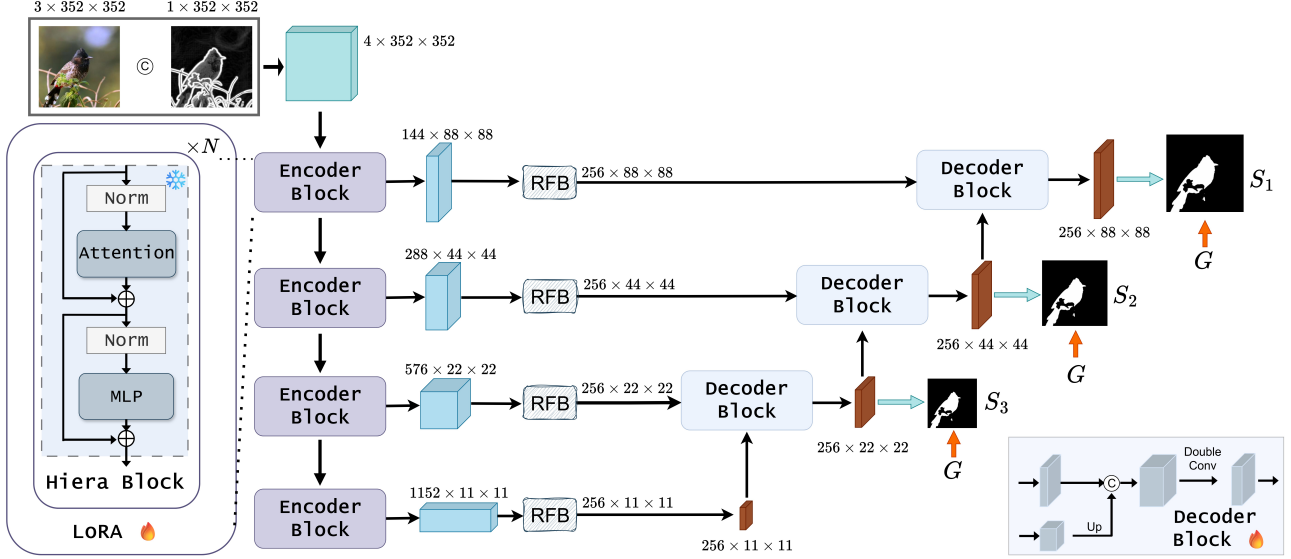}
    \caption{Overview of the proposed BED-SAM2 architecture, adapted from SAM2-UNet~\cite{xiong2026sam2_unet}. A LoRA-augmented Hiera encoder extracts multi-scale features, which are refined and decoded in a U-Net–style framework to produce segmentation predictions.}
    \label{fig:BED-SAM2}
\end{figure*}

Some COD-focused work has explored implementing MDE. Earlier studies, written when MDE techniques were significantly weaker, assumed that monocular depth maps were insufficiently accurate. Xiang et al.~\cite{xiang2022exploringdepthcontributioncamouflaged_mdeCOD} introduced a confidence-aware mechanism that uses a generative adversarial network (GAN) to estimate the trustworthiness of each depth pixel. Depth-aided COD (DaCOD)~\cite{10.1145/3581783.3611874_DaCOD} uses a hybrid CNN+Transformer approach that processes RGB with a ResNet backbone for local features and depth with a SwinIR backbone to capture global features. 

In summary, while prior works improve SOD with either VFMs or depth-assisted networks, none achieve lightweight, early-fusion integration of depth cues into a pre-trained VFM. BED-SAM2 addresses this gap by introducing a simple early-fusion strategy that feeds monocular depth-derived edges directly into the SAM2 Hiera encoder and fine-tunes using LoRA-based adaptation. Our method enables SAM2 to encode geometric boundary cues without additional backbones, depth sensors, or complex fusion modules.

\section{Methodology}

Our proposed method, BED-SAM2 (as shown in Fig.~\ref{fig:BED-SAM2}), takes a straightforward approach. We extend the SAM2 Hiera encoder by introducing a fourth input channel for edge data calculated from monocular depth; input-level early fusion allows the SAM2 encoder to process RGB and depth edge information in parallel. The edge channel weights are initialized by taking the mean of the pretrained RGB channel weights from SAM2's encoder. The addition of depth edges provides detailed visual cues and strong boundaries for segmentation, and demonstrates the extendability of the SAM2 encoder to quickly learn unseen modalities.

\subsection{Encoder}
Given an input 
$\textit{I} \in \mathbb{R} ^ {(4\times H \times W)}$ 
where \textit{H} denotes height and \textit{W} denotes width, 
The SAM2 Hiera encoder outputs four hierarchical feature levels:
\begin{equation}
\begin{split}
    X_i &\in \mathbb{R}^{C_i \times \frac{H}{2^{i+1}} \times \frac{W}{2^{i+1}}}, \\
    i &\in \{1, 2, 3, 4\}, \quad C_i \in \{144, 288, 576, 1152\}
\end{split}
\end{equation}

We inject every layer of our encoder with trainable LoRA~\cite{hu2021loralowrankadaptationlarge} rank decomposition matrices. To accommodate the additional geometric prior, we modify the first layer of the Hiera encoder (the patch embedding layer) to accept a 4-channel input. Let $W \in \mathbb{R}^{C_{out} \times 3 \times k \times k}$ represent the pre-trained weights of the first convolution. We initialize the expanded weight tensor $W' \in \mathbb{R}^{C_{out} \times 4 \times k \times k}$ such that the first three channels are copied from $W$, and the fourth channel $W'_4$ is initialized as the channel-wise mean: $W'_4 = \frac{1}{3}\sum_{j=1}^3 W_j$. This ensures that the depth-derived edges are processed within the same feature scale as the RGB data at the onset of training

\subsection{Receptive Field Blocks}
Following the work of SAM2-UNet~\cite{xiong2026sam2_unet}, the encoded features are passed through four receptive field blocks (RFBs)~\cite{liu2018receptivefieldblocknet,fan2020pranetparallelreverseattention_structure_loss} to reduce the number of feature channels. We use RFBs of size 256 (instead of 64 in SAM2-UNet) to provide a larger receptive field. This allows the network to scale for larger input images (e.g., $1024 \times 1024$ instead of $352 \times 352$) in high-resolution segmentation.

\subsection{U-shaped Decoder}
The mask decoder of SAM2 consists of a two-way transformer between the prompt embedding and encoder features~\cite{ravi2024sam2segmentimages}. We instead use a classic UNet decoder~\cite{ronneberger2015unetconvolutionalnetworksbiomedical_unet} that consists of three decoder blocks between feature levels that upsample and combine the hierarchical features at each level. Bilinear upsampling is used on lower resolution features, concatenated with the level above it, then passed through a double convolution block consisting of two sequential ``Conv-BN-ReLu'' operations at each level, where ``Conv'' is a $3 \times 3$ convolution layer and ``BN'' is batch normalization. Finally, each feature is passed through a $1 \times 1$ convolutional segmentation head to produce binary segmentation masks $\textit{S}_i (i \in 1,2,3)$ supervised by ground truth \textit{G}.

\subsection{Loss Function}
We adopt the structure loss used in~\cite{fan2020pranetparallelreverseattention_structure_loss,Wei_Wang_Huang_2020_structure_loss,xiong2026sam2_unet} which combines weighted binary cross-entropy (BCE) and weighted IoU to emphasize structural consistency along object boundaries: 
\begin{equation}
    \mathcal{L} = \mathcal{L}_{IoU}^w + \mathcal{L}_{BCE}^w
\end{equation}
following~\cite{xiong2026sam2_unet}, this loss is applied for each level in the segmentation output; the total loss is formulated as: 
\begin{equation}
    \mathcal{L}_{total} = \sum_{i=1}^3\mathcal{L}(G, S_i)
\end{equation}

\subsection{Calculating Depth Edges}
First, we obtain our depth images using Distill any Depth~\cite{he2025distilldepthdistillationcreates}. The depth information contains less noise than RGB and emphasizes object boundaries over color and texture. This depth information helps to address a significant problem in existing SOD models: establishing strong object boundaries. 

To extract the most information from our depth, we 
calculate edges for the given depth, inverse depth, and then 
perform a depth enhancement to emphasize objects at middle depth values. The depth enhancement is defined as:

\begin{equation}
D'(x,y) = \left| D(x,y) - 0.5 \right| \cdot 2
\end{equation}

where \textit{D} $\in [0,1]$ is the normalized depth and \textit{D'} is the enhanced depth. This transformation sets 0.5 (mid-depth) to zero and maps values farther from 0.5 toward 1, highlighting objects at middle depth. We then pass the three depth images through a classic Sobel operator~\cite{kanopoulos1988design_Sobel} to extract soft edges. By using soft edges with a gradient $E(x,y) \in [0,1]$ over binary strong edges $E(x,y) \in \{0,1\}$, we allow the soft edges between depth images to compound and be weighted more heavily than edges that might only appear in one or two of the depth images (e.g., In Fig. \ref{fig:bird_depth} the wing on the left side is separated from the hummingbird because there is a strong edge separating the body and wing.)

The final structure map $S_{map}$ is obtained by the element-wise 
summation of the soft edges derived from each depth representation:
\begin{equation}
    S_{map} = \mathcal{S}(D) + \mathcal{S}(D_{inv}) + \mathcal{S}(D')
\end{equation}
where $\mathcal{S}$ denotes the Sobel operator. By aggregating these 
gradients, $S_{map}$ emphasizes consistent structural discontinuities 
while suppressing transient depth noise.

\section{Experiments}

\subsection{Experimental Setup}
\noindent\textbf{Datasets.} BED-SAM2 is evaluated on thirteen benchmark datasets spanning three distinct tasks:
\begin{itemize}
    \item \textbf{Salient Object Detection (SOD):} Trained on DUTS-TR~\cite{wang2017_DUTS} and subsets of high-resolution datasets HRSOD~\cite{zeng2019highresolutionsalientobjectdetection_HRSOD}, and UHRSD~\cite{xie2022pyramid_pgnet_uhrsd}. Evaluated on five standard benchmarks: DUTS-TE, ECSSD~\cite{Yan_2013_CVPR_ECSSD}, DUT-OMRON~\cite{yang2013saliency_DUT-OMRON}, HKU-IS~\cite{li2015visualsaliencybasedmultiscale_HKU-IS}, and PASCAL-S~\cite{li2014secretssalientobjectsegmentation_PASCAL-S}.

    \item \textbf{Depth-Guided Salient Object Detection (RGBD-SOD):}
    Provides ground truth depth in addition to RGB data. Trained on a subset of NJU2K~\cite{NJU2K} and NLPR~\cite{NLPR}. Evaluated on four standard benchmarks: NJU2K, NLPR, SIP~\cite{Fan_2021_SIP}, and STERE~\cite{6247708_STERE}.
    
    \item \textbf{Camouflaged Object Detection (COD):} Trained on subsets of CAMO~\cite{ltnghia-CVIU2019_CAMO} and COD10K~\cite{Fan_2020_CVPR_COD10K}. Evaluation is performed on CAMO, COD10K, CHAMELEON~\cite{skurowski2018animal_CHAMELEON}, and NC4K~\cite{yunqiu_cod21_nc4k}.
\end{itemize}

\subsection{Metrics}
We utilize four metrics for comparison to state-of-the-art models and benchmarks. S-measure (\textit{$S_m$})~\cite{fan2017structuremeasurenewwayevaluate} , F-measure (max,weighted) \textit{$(F_\beta^x, F_\beta^w)$}~\cite{Margolin_2014_CVPR_F-measure}, mean E-measure  \textit{$(E_\phi)$}~\cite{fan2018enhancedalignmentmeasurebinaryforeground}, and mean absolute error (MAE).

\begin{itemize}
\item S-measure (structure-measure, \textit{$S_m$})~\cite{fan2017structuremeasurenewwayevaluate} measures the structural similarity between the predicted saliency and ground truth masks. The formula for S-measure is derived as:

\begin{equation}
    S_\alpha = \alpha \cdot S_o + (1-\alpha) \cdot S_r
\end{equation}

where $\alpha \in [0,1]$ and by default is 0.5, $S_o$ and $S_r$ denote object aware and region-aware structural similarity respectively.

\vspace{10px}

\item F-measure \textit{$(F_\beta)$}~\cite{Margolin_2014_CVPR_F-measure} is used to combine the different qualities of the binary map, true-positive \textit{TP}, true-negative \textit{TN}, false-positive \textit{FP}, and false-negative \textit{FN}. \textit{$F_\beta$} measure is derived as:

\begin{equation}
    F_\beta =  \frac{(1+\beta^2)(Precision \cdot Recall)}{(\beta^2 \cdot Precision + Recall)}  
\end{equation}

where $\beta$ is a hyperparameter that controls the preference between a confident and complete detection vs over-detection, the default value for $\beta^2$ in SOD is typically $0.3$ to emphasize precision over recall~\cite{fan2017structuremeasurenewwayevaluate}. 

\vspace{10px}

\item E-measure (enhanced alignment measure, \textit{$E_\phi$})~\cite{fan2018enhancedalignmentmeasurebinaryforeground} captures both local pixel-level and global image-level properties between saliency and ground truth masks. E-measure is defined as:

\begin{equation}
    E_\phi = \frac{1}{w \cdot h}\sum_{x=1}^w\sum_{y=1}^hM_\phi(x,y)
\end{equation}

where \textit{h} and \textit{w} are the height and width of the masks, $M$ is the enhanced alignment matrix, and $\phi$ is the foreground map. We adopt the mean E-measure for our evaluation ($E_\phi^m)$.

\end{itemize}

Evaluation across all these performance metrics is essential for quantifying the success of an SOD network and identifying where a model might struggle to generalize.

\subsubsection{RGB Salient Object Detection}
Most state-of-the-art models traditionally train on DUTS-TR. We examine the use of additional training data with high-resolution datasets HRSOD~\cite{zeng2019highresolutionsalientobjectdetection_HRSOD} and UHRSD~\cite{xie2022pyramid_pgnet_uhrsd}, as they have previously been shown to improve results on several of our key test datasets~\cite{Zheng_2024_BiRefNet}. BED-SAM2 is evaluated on DUTS-TE, DUT-OMRON, ECSSD, HKU-IS, and PASCAL-S. Models are trained for three different input resolutions ($352 \times 352$, $512 \times 512$, and $1024 \times 1024$) for direct comparisons to various state-of-the-art methods.

\subsection{RGB-D Salient Object Detection}
Four popular RGB-D benchmark datasets provide depth information. NJU2K~\cite{NJU2K} is a stereo dataset collected from the internet and 3D movies. STERE~\cite{6247708_STERE} is a stereo dataset collected from the internet. SIP~\cite{Fan_2021_SIP} is a salient person dataset collected with a Huawei Mate 10 that aims to capture human poses in various environments, and NLPR~\cite{NLPR} is an indoor/outdoor dataset collected using a Microsoft Kinect for LiDAR depth.

\subsection{Camouflaged Object Detection}
Camouflaged object detection (COD) introduced by~\cite{Fan_2022_COD} seeks to detect objects that blend in with their surroundings. 
For camouflaged object detection, we follow~\cite{Zheng_2024_BiRefNet,xiong2026sam2_unet,Fan_2022_COD}, training BED-SAM2 on a combination of the CAMO-TR~\cite{ltnghia-CVIU2019_CAMO} and COD10K-TR~\cite{Fan_2020_CVPR_COD10K} datasets. We evaluate across CAMO-TE~\cite{ltnghia-CVIU2019_CAMO}, COD10K-TE~\cite{Fan_2020_CVPR_COD10K}, CHAMELEON~\cite{skurowski2018animal_CHAMELEON}, and NC4K~\cite{yunqiu_cod21_nc4k}. As these datasets intentionally seek to deceive human perception, they do not offer as much trustworthy information through RGB as they do with true depth information. 

\subsection{Implementation}
Our model is trained on a single RTX 5090 with 32GB VRAM, 64GB RAM, and an AMD 9950X3D. We use the AdamW optimizer~\cite{loshchilov2019decoupledweightdecayregularization_AdamW} with a weight decay of $5e^{-4}$. For our high-resolution and low-resolution salient object detection models, we train over $5$ epochs with a learning rate of $1e^{-4}$ and a cosine decay of $5e^{-4}$. Depending on the size of our input images, $352 \times 352$, $512 \times 512$, and $1024 \times 1024$, we use batch sizes of $24$, $12$, and $3$, respectively. Data augmentation consists of random vertical and horizontal flips, and we use the latest SAM2.1 Hiera-L checkpoint.

For COD, we train over $30$ epochs with a learning rate of $5e^{-4}$, all other parameters remain the same.


BED-SAM2 fine-tuning converges significantly faster than current state-of-the-art methods. For comparison, BiRefNet is trained for $150$ epochs on eight A100 GPUs, and the original SAM2-UNet trains for $50$ epochs on a single RTX 4090.

\subsection{Ablation}
Table \ref{tab:Ablation_Results} demonstrates the efficacy of our small architectural changes, we perform an ablation on the DUTS-TE dataset to examine our use of LoRA~\cite{hu2021loralowrankadaptationlarge} instead of SAM2-Adapter~\cite{chen2024sam2adapterevaluatingadapting}, increasing the size of the RFB from 64 to 256, and the use of depth information vs edge information. We perform a separate ablation in Table \ref{tab:Ablation_CumulativeMap} to justify the benefit of our novel cumulative structure map.

\begin{table}[htbp]
\centering
\caption{Effects of architecture components on BED-SAM2 evaluated on the DUTS-TE dataset. All models are trained exclusively on DUTS-TR with an input resolution of $352$.}
\label{tab:Ablation_Results}

\setlength{\tabcolsep}{3.5pt}

\resizebox{\columnwidth}{!}{%
\scriptsize
\begin{tabular}{l cccc cccc} 
\toprule
\multirow{2}{*}{Methods} 
& \multicolumn{4}{c}{Options} 
& \multicolumn{4}{c}{\textbf{DUTS-TE (5019)}} \\
\cmidrule(lr){2-5} \cmidrule(lr){6-9} 
& Depth & Edges & LoRA & RFB 
& $S_m \uparrow$ & $F_\beta^{x} \uparrow$ & $E_\phi \uparrow$ & $\mathcal{M} \downarrow$ \\
\midrule

 & \checkmark & & & 
& .922 & .918 & .944 & .028 \\

 & & \checkmark & & 
& .926 & .920 & .945 & .026 \\

 & & & \checkmark & 
& .926 & .926 & .946 & .027 \\

 & & & & \checkmark 
& .933 & .928 & .954 & .022 \\


 & \checkmark & & \checkmark & 
& .930 & .927 & .949 & .026 \\

BED-SAM2 & \checkmark & & & \checkmark 
& .933 & .929 & .955 & .022 \\

 & & \checkmark & \checkmark & 
& .931 & .927 & .950 & .025 \\

 & & \checkmark & & \checkmark 
& .934 & .930 & .956 & .021 \\

 & \checkmark & & \checkmark & \checkmark 
& .936 & .933 & .957 & .021 \\

 & & \checkmark & \checkmark & \checkmark 
& \textbf{.937} & \textbf{.934} & \textbf{.957} & \textbf{.021} \\

\bottomrule
\end{tabular}%
} 
\end{table}
\begin{table}[htbp]
\centering
\caption{Ablation for sobel edge structure maps. The model was trained with an input resolution of 352 on DUTS-TR and evaluated against CAMO.}
\label{tab:Ablation_CumulativeMap}

\setlength{\tabcolsep}{3.5pt}

\resizebox{\columnwidth}{!}{%

\begin{tabular}{l cccc} 
\toprule
\multirow{2}{*}{Methods} 
& \multicolumn{4}{c}{\textbf{CAMO}} \\
\cmidrule(lr){2-5} 
& $S_m \uparrow$ & $F_\beta^{x} \uparrow$ & $E_\phi \uparrow$ & $\mathcal{M} \downarrow$ \\
\midrule

Depth Edges
& .758 & .628 & .766 & .094 \\

Inverted Depth Edges
& .760 & .632 & .773 & .094 \\

Centered Depth Edges
& .768 & .641 & .780 & .091 \\

Cumulative Edges
& \textbf{.772} & \textbf{.648} & \textbf{.783} & \textbf{.090} \\

\bottomrule
\end{tabular}%
} 
\end{table}

\section{Results}

\begin{table*}[htbp]
\centering
\caption{BED-SAM2 trained at different resolutions and evaluated on five RGB SOD datasets. Size denotes the training resolution. TR denotes the datasets used for training, where 1, 2, and 3 represent DUTS-TR~\cite{wang2017_DUTS}, HRSOD-TR~\cite{zeng2019highresolutionsalientobjectdetection_HRSOD}, and UHRSD-TR~\cite{xie2022pyramid_pgnet_uhrsd}, respectively. Best results are highlighted in green, second best in blue.}
\label{tab:SOD_Results}

\resizebox{\textwidth}{!}{%
\scriptsize 
\begin{tabular}{@{} l c c cccc cccc cccc cccc cccc @{}} 
\toprule
\multirow{2}{*}{Methods} & \multirow{2}{*}{Size} & \multirow{2}{*}{TR} 
& \multicolumn{4}{c}{\textbf{DUTS-TE (5019)}} 
& \multicolumn{4}{c}{\textbf{DUT-OMRON (5168)}} 
& \multicolumn{4}{c}{\textbf{ECSSD (1000)}} 
& \multicolumn{4}{c}{\textbf{HKU-IS (4447)}} 
& \multicolumn{4}{c}{\textbf{PASCAL-S (850)}} \\
\cmidrule(lr){4-7} \cmidrule(lr){8-11} \cmidrule(lr){12-15} \cmidrule(lr){16-19} \cmidrule(lr){20-23}
& & 
& $S_m \uparrow$ & $F_\beta^{x} \uparrow$ & $E_\phi^m \uparrow$ & $\mathcal{M} \downarrow$ 
& $S_m \uparrow$ & $F_\beta^{x} \uparrow$ & $E_\phi^m \uparrow$ & $\mathcal{M} \downarrow$ 
& $S_m \uparrow$ & $F_\beta^{x} \uparrow$ & $E_\phi^m \uparrow$ & $\mathcal{M} \downarrow$
& $S_m \uparrow$ & $F_\beta^{x} \uparrow$ & $E_\phi^m \uparrow$ & $\mathcal{M} \downarrow$
& $S_m \uparrow$ & $F_\beta^{x} \uparrow$ & $E_\phi^m \uparrow$ & $\mathcal{M} \downarrow$ \\
\midrule
MENet$_{23}$~\cite{Wang_2023_CVPR_MENet} & 352 & 1 
& .905 & .912 & .937 & .028
& .850 & .834 & .891 & .045
& .928 & .955 & .954 & .031
& .927 & .948 & .966 & .023
& .872 & .890 & .913 & .054\\
EDN$_{22}$~\cite{Wu_2022_EDN} & 384 & 1 
& .893 & .844 & - & .035
& .821 & .770 & - & .050
& .950 & .918 & - & .033
& .940 & .908 & - & .027
& .879 & .827 & - & .062\\
ICON-S$_{21}$~\cite{zhuge2022salientobjectdetectionintegrity_icon} & 352 & 1 
& .917 & .886 & .954 & .025
& .869 & .804 & .900 & .043
& .941 & .936 & .966 & .023
& .935 & .925 & .968 & .022
& .885 & .854 & .924 & .048\\
U2Net$_{20}$~\cite{Qin_2020_U2Net} & 320 & 1 
& .861 & .873 & - & .044
& .847 & .823 & - & .054
& .928 & .951 & - & .033
& .916 & .935 & - & .031
& .844 & .859 & - & .074\\
PGNet$_{22}$~\cite{xie2022pyramid_pgnet_uhrsd} & 1024   & 1 
& .911 & .917 & .922 & .027
& .855 & .835 & .887 & .045
& - & - & - & -
& - & - & - & -
& - & - & - & -\\
PGNet$_{22}$~\cite{xie2022pyramid_pgnet_uhrsd} & 1024   & [1,2] 
& .912 & .919 & .925 & .028
& .858 & .835 & .887 & .046
& - & - & - & -
& - & - & - & -
& - & - & - & -\\
PGNet$_{22}$~\cite{xie2022pyramid_pgnet_uhrsd} & 1024   & [2,3] 
& .859 & .871 & .897 & .038
& .786 & .772 & .884 & .058
& - & - & - & -
& - & - & - & -
& - & - & - & -\\
VSCode$_{23}$~\cite{luo2024vscodegeneralvisualsalient} & 352   & 1 
& .926 & .922 & .960 & -
& .877 & .840 & .912 & -
& .949 & .959 & \best{.974} & -
& .940 & .951 & \best{.974} & -
& .887 & .864 & .904 & - \\ 
BiRefNet$_{24}$~\cite{Zheng_2024_BiRefNet} & 1024   & 1 
& .939 & .937 & .958 & .019
& .868 & .813 & .878 & .040
& - & - & - & -
& - & - & - & -
& - & - & - & -\\
BiRefNet$_{24}$~\cite{Zheng_2024_BiRefNet} & 1024   & [1,2]
& .938 & .935 & .960 & \second{.018}
& .868 & .818 & .882 & .040
& - & - & - & -
& - & - & - & -
& - & - & - & -\\
BiRefNet$_{24}$~\cite{Zheng_2024_BiRefNet} & 1024   & [1,3]
& .942 & .942 & .961 & \second{.018}
& .881 & .837 & .896 & .036
& - & - & - & -
& - & - & - & -
& - & - & - & -\\
BiRefNet$_{24}$~\cite{Zheng_2024_BiRefNet} & 1024   & [2,3]
& .933 & .928 & .954 & .020
& .864 & .810 & .879 & .040
& - & - & - & -
& - & - & - & -
& - & - & - & -\\
BiRefNet$_{24}$~\cite{Zheng_2024_BiRefNet} & 1024   & [1,2,3]
& \best{.944} & .943 & .962 & \second{.018}
& .882 & .839 & .896 & .038
& - & - & - & -
& - & - & - & -
& - & - & - & -\\
SAM2-UNet$_{26}$~\cite{xiong2026sam2_unet} & 352 & 1 
& .934 & - & .959 & .020
& .884 & - & .912 & .039
& .950 & - & .970 & .020
& .941 & - & \second{.971} & \second{.019}
& .894 & - & .931 & .043\\
\midrule
\rowcolor{samgray}
BED-SAM2 & 352   & 1 
& .937 & .933 & .959 & .021 
& .886 & .849 & .908 & .038 
& .951 & .962 & .966 & .022 
& \second{.946} & .954 & .968 & .020 
& .894 & .894 & .923 & .047\\
\rowcolor{samgray}
BED-SAM2 & 352   & [1,2]
& .939 & .938 & .959 & .020 
& .891 & .858 & .912 & .036 
& .953 & .964 & .967 & .021 
& .944 & .954 & .966 & .021 
& \second{.898} & .896 & .930 & .044\\
\rowcolor{samgray}
BED-SAM2 & 352  & [1,3] 
& .940 & .939 & .961 & .020 
& \second{.897} & .867 & .918 & \second{.034} 
& \best{.955} & .966 & .969 & .020 
& \second{.946} & .955 & .967 & .020 
& .896 & .894 & .927 & .045\\
\rowcolor{samgray}
BED-SAM2 & 352   & [2,3]
& .930 & .927 & .950 & .024 
& .887 & .848 & .903 & .039 
& .951 & .961 & .964 & .023 
& .934 & .945 & .955 & .026 
& .895 & .891 & .926 & .046\\
\rowcolor{samgray}
BED-SAM2 & 352   & [1,2,3]
& .941 & .941 & .962 & .020 
& \best{.898} & \best{.871} & \second{.921} & .035 
& \best{.955} & \second{.967} & .970 & .019 
& .945 & \second{.956} & .967 & .020 
& \second{.898} & .898 & .931 & .043\\
\midrule
\rowcolor{samgray}
BED-SAM2 & 512   & 1 
& \second{.943} & .942 & .963 & \second{.018} 
& .891 & .860 & .915 & .035 
& .952 & .963 & .966 & .021
& \best{.947} & \best{.957} & \second{.971} & \best{.018} 
& \second{.898} & \second{.904} & .930 & .043\\
\rowcolor{samgray}
BED-SAM2 & 512   & [1,2] 
& \second{.943} & .943 & .963 & \second{.018}
& .891 & .858 & .914 & .035
& .952 & .963 & .967 & .020
& \second{.946} & \best{.957} & .970 & \second{.019}
& \best{.901} & .899 & \second{.934} & \best{.041}   \\
\rowcolor{samgray}
BED-SAM2 & 512  & [1,3]
& \best{.944} & .942 & .961 & .019
& \second{.897} & .868 & .918 & \second{.034}
& \best{.955} & .966 & .969 & \second{.018}
& \best{.947} & \second{.956} & .968 & .020
& .897 & .897 & .930 & .043   \\
\rowcolor{samgray}
BED-SAM2 & 512   & [2,3] 
& .934 & .934 & .954 & .022 
& .888 & .857 & .911 & .037 
& \second{.954} & .965 & .968 & .019 
& .936 & .948 & .960 & .024 
& .894 & .895 & .929 & .044\\
\rowcolor{samgray}
BED-SAM2 & 512   & [1,2,3] 
& \best{.944} & \second{.946} & \best{.965} & \best{.017} 
& .896 & \second{.869} & \best{.922} & \second{.034} 
& .953 & .966 & .969 & \second{.018} 
& .944 & \second{.956} & .969 & \second{.019} 
& \second{.898} & .900 & \second{.934} & \best{.041}\\
\midrule
\rowcolor{samgray}
BED-SAM2 & 1024   & 1 
& .941 & .940 & .960 & \second{.018}
& .882 & .842 & .899 & .035
& \best{.955} & \best{.968} & \second{.971} & \best{.017}
& .945 & \best{.957} & .970 & \best{.018}
& \second{.898} & \best{.906} & \second{.934} & \second{.042}\\
\rowcolor{samgray}
BED-SAM2 & 1024   & [1,2]
& \second{.943} & .945 & .963 & \best{.017}
& .889 & .854 & .911 & \second{.034}
& \second{.954} & \second{.967} & .970 & \second{.018}
& .943 & .955 & .970 & \second{.019}
& .897 & .898 & \best{.935} & \best{.041}\\
\rowcolor{samgray}
BED-SAM2 & 1024  & [1,3] 
& .942 & .944 & .963 & \second{.018}
& .893 & .861 & .914 & \best{.032}
& .953 & .965 & .968 & \second{.018} 
& .943 & \second{.956} & .969 & \second{.019}
& .891 & .890 & .930 & .043\\
\rowcolor{samgray}
BED-SAM2 & 1024   & [2,3] 
& .926 & .927 & .949 & .024 
& .876 & .836 & .897 & .040 
& .952 & .965 & .969 & .019 
& .924 & .939 & .954 & .027 
& .887 & .890 & .926 & .047 \\
\rowcolor{samgray}
BED-SAM2 & 1024   & [1,2,3] 
& \best{.944} & \best{.947} & \second{.964} & \best{.017} 
& .891 & .857 & .912 & \second{.034} 
& \second{.954} & .965 & .968 & \second{.018} 
& .942 & .954 & .968 & \second{.019} 
& .895 & .896 & .932 & \second{.042} \\
\bottomrule
\end{tabular}%
} 
\end{table*}
\begin{table*}[htbp]
\centering
\caption{Evaluation of models trained across various input resolutions and depth modalities. The models were evaluated on four RGB-D SOD datasets. “Size” denotes the training resolution. “TR” denotes the datasets used for training, where 4, and 5 represent NJU2K-TR, and NLPR-TR, respectively. Variations include training on RGB paired with original depth (D), original depth edges (E), or monocular depth (MD). Best results are highlighted in green, second best in blue.}
\label{tab:RGBD_Results}

\setlength{\tabcolsep}{3.5pt} 
\renewcommand{\arraystretch}{0.95} 

\resizebox{\textwidth}{!}{%
\scriptsize 
\begin{tabular}{@{} l c c cccc cccc cccc cccc @{}} 
\toprule
\multirow{2}{*}{Methods} & \multirow{2}{*}{Size} & \multirow{2}{*}{TR} 
& \multicolumn{4}{c}{\textbf{NJU2K (500)}} 
& \multicolumn{4}{c}{\textbf{NLPR (300)}} 
& \multicolumn{4}{c}{\textbf{SIP (929)}} 
& \multicolumn{4}{c}{\textbf{STERE (1000)}} \\
\cmidrule(lr){4-7} \cmidrule(lr){8-11} \cmidrule(lr){12-15} \cmidrule(lr){16-19}
& & 
& $S_m \uparrow$ & $F_\beta^{x} \uparrow$ & $E_\phi^m \uparrow$ & $\mathcal{M} \downarrow$ 
& $S_m \uparrow$ & $F_\beta^{x} \uparrow$ & $E_\phi^m \uparrow$ & $\mathcal{M} \downarrow$ 
& $S_m \uparrow$ & $F_\beta^{x} \uparrow$ & $E_\phi^m \uparrow$ & $\mathcal{M} \downarrow$
& $S_m \uparrow$ & $F_\beta^{x} \uparrow$ & $E_\phi^m \uparrow$ & $\mathcal{M} \downarrow$ \\
\midrule
MonoSOD$_{21}$ \cite{9561211_monoSOD} & 224 &  [4,5]
& .886 & .877 & .917 & .049 
& .904 & .878 & .944 & .030 
& - & - & - & - 
& .889 & .871 & .921 & .047  \\ 
CMINet$_{21}$ \cite{zhang2022rgbdsaliencydetectioncascaded} & 352 &  [4,5]
& \best{.939} & .925 & \second{.956} & .032 
& \second{.941} & .909 & .964 & .019 
& .894 & .887 & .933 & .044
& - & - & - & -  \\ 
HiDAnet$_{23}$ \cite{Wu_2023_hidanet} & 352 &  [4,5]
& .926 & \second{.939} & .954 & \second{.029}
& .930 & .929 & .961 & .021
& .892 & .919 & .927 & .043
& .911 & .921 & .946 & .035  \\ 
FasterSal$_{25}$ \cite{10814716_fastersal} & 256 &  [4,5]
& .908 & .906 & .949 & .034
& .920 & .902 & .960 & .022
& .870 & .870 & .929 & .049
& .888 & .875 & .939 & .040  \\ 
DFormerV2$_{25}$ \cite{dformerv2} & 480x640 &  [4,5]
& \second{.937} & \best{.946} & \best{.964} & \best{.023} 
& \best{.942} & \second{.939} & .971 & .016 
& .915 & .938 & .950 & .032 
& .923 & .929 & .952 & .030 \\ 
\midrule
BED-SAM2 (D) & 352   & [4,5] 
& .917 & .917 & .929 & .042
& .934 & .932 & .958 & .017
& .920 & .934 & .938 & .036
& .931 & .925 & .943 & .032\\
BED-SAM2 (E) & 352   & [4,5] 
& .918 & .916 & .929 & .041
& .931 & .928 & .958 & .017
& .917 & .931 & .936 & .036
& .930 & .924 & .943 & .033\\
BED-SAM2 (MD) & 352   & [4,5] 
& .920 & .922 & .933 & .041
& .934 & .932 & .958 & .017
& .918 & .934 & .936 & .037
& .932 & .926 & .945 & .031\\
\midrule
BED-SAM2 (D) & 512   & [4,5] 
& .920 & .919 & .929 & .041
& .937 & .934 & .959 & .016
& .920 & .931 & .935 & .036
& \second{.935} & .930 & .944 & .031\\
BED-SAM2 (E) & 512   & [4,5] 
& .920 & .915 & .929 & .040
& \second{.941} & .936 & .964 & \second{.014}
& .918 & .932 & .938 & .035
& .934 & .930 & .947 & .030\\
BED-SAM2 (MD) & 512   & [4,5] 
& .922 & .919 & .931 & .040
& .939 & .935 & .961 & .015
& .919 & .932 & .936 & .036
& .934 & .931 & .945 & .031\\
\midrule
BED-SAM2 (D) & 1024   & [4,5] 
& .924 & .925 & .939 & .035
& .939 & \second{.939} & .964 & \second{.014}
& .923 & .939 & .946 & .030
& \best{.937} & \best{.935} & .954 & \best{.026}\\
BED-SAM2 (E) & 1024   & [4,5] 
& .922 & .921 & .937 & .037
& .940 & \second{.939} & .967 & \second{.014}
& .919 & .940 & .946 & .031
& .934 & .933 & .952 & \best{.026}\\
BED-SAM2 (MD) & 1024  & [4,5] 
& .925 & .926 & .939 & .036
& .937 & .937 & .964 & \second{.014}
& .924 & .943 & .950 & .029
& .933 & \second{.934} & .952 & \second{.027}\\
\midrule
\rowcolor{samgray}
BED-SAM2 & 352   & 1 
& .900 & .895 & .929 & .049 
& .911 & .901 & .948 & .028 
& .930 & .936 & .959 & .027 
& .927 & .925 & .956 & .030  \\
\rowcolor{samgray}
BED-SAM2 & 352   & [1,2,3]
& .910 & .908 & .937 & .045 
& .920 & .910 & .954 & .024 
& \second{.941} & \second{.957} & \second{.971} & .022
& .928 & .931 & .958 & .031 \\
\rowcolor{samgray}
BED-SAM2 & 352   & [4,5]
& .917 & .917 & .942 & .042 
& .940 & .937 & \best{.973} & .015 
& .913 & .930 & .946 & .037 
& .934 & .929 & \second{.960} & .029 \\
\midrule
\rowcolor{samgray}
BED-SAM2 & 512   & 1 
& .904 & .903 & .933 & .047 
& .914 & .896 & .942 & .027 
& .934 & .942 & .963 & .025 
& .927 & .928 & .957 & .030 \\
\rowcolor{samgray}
BED-SAM2 & 512   & [1,2,3] 
& .910 & .908 & .938 & .043 
& .911 & .907 & .947 & .027 
& .940 & \second{.957} & .970 & \second{.021} 
& .929 & .933 & \second{.960} & .029 \\
\rowcolor{samgray}
BED-SAM2 & 512   & [4,5]
& .923 & .918 & .940 & .039 
& .940 & .937 & \best{.973} & .015 
& .919 & .930 & .949 & .036 
& .931 & .929 & .959 & .032 \\
\midrule
\rowcolor{samgray}
BED-SAM2 & 1024   & 1 
& .902 & .898 & .930 & .048 
& .911 & .901 & .945 & .026 
& .937 & .948 & .965 & .023 
& .928 & .927 & .957 & .028 \\
\rowcolor{samgray}
BED-SAM2 & 1024   & [1,2,3] 
& .911 & .913 & .939 & .043 
& .915 & .909 & .952 & .026 
& \best{.945} & \best{.962} & \best{.974} & \best{.019} 
& .931 & \second{.934} & \best{.961} & \second{.027} \\
\rowcolor{samgray}
BED-SAM2 & 1024   & [4,5]
& .923 & .924 & .944 & .037 
& \best{.942} & \best{.941} & \second{.972} & \best{.013} 
& .923 & .942 & .955 & .030 
& \second{.935} & .933 & .959 & \second{.027} \\
\bottomrule
\end{tabular}%
} 
\end{table*}
\begin{table*}[htbp]
\centering
\caption{Quantitative comparisons between our models trained at different resolutions and evaluated on four COD datasets. Size denotes the training resolution. Best results are highlighted in green, second best in blue.}
\label{tab:COD_Results}

\setlength{\tabcolsep}{4pt} 
\renewcommand{\arraystretch}{0.95} 

\resizebox{\textwidth}{!}{%
\scriptsize 
\begin{tabular}{@{} l c cccc cccc cccc cccc @{}} 
\toprule
\multirow{2}{*}{Methods} & \multirow{2}{*}{Size} 
& \multicolumn{4}{c}{\textbf{CAMO (250)}} 
& \multicolumn{4}{c}{\textbf{COD10K (2026)}} 
& \multicolumn{4}{c}{\textbf{CHAMELEON (76)}} 
& \multicolumn{4}{c}{\textbf{NC4K (4121)}} \\
\cmidrule(lr){3-6} \cmidrule(lr){7-10} \cmidrule(lr){11-14} \cmidrule(lr){15-18}
&
& $S_m \uparrow$ & $F_\beta^{w} \uparrow$ & $E_\phi^m \uparrow$ & $\mathcal{M} \downarrow$ 
& $S_m \uparrow$ & $F_\beta^{w} \uparrow$ & $E_\phi^m \uparrow$ & $\mathcal{M} \downarrow$ 
& $S_m \uparrow$ & $F_\beta^{w} \uparrow$ & $E_\phi^m \uparrow$ & $\mathcal{M} \downarrow$
& $S_m \uparrow$ & $F_\beta^{w} \uparrow$ & $E_\phi^m \uparrow$ & $\mathcal{M} \downarrow$ \\
\midrule
ZoomNeXt$_{23}$ \cite{Pang_2024_Zoomnext} & 384  
& .889 & .857 & \second{.945} & .041
& .898 & .827 & \second{.956} & .018
& .924 & .885 & \best{.975} & \second{.018}
& .903 & .863 & \second{.951} & \second{.028}\\
VSCode$_{24}$ \cite{luo2024vscodegeneralvisualsalient} & 352 
& .873 & .861 & .938 & -
& .869 & .827 & .942 & -
& - & - & - & -
& .891 & \second{.878} & .944 & -\\

BiRefNet$_{24}$ \cite{Zheng_2024_BiRefNet} & 1024  
& \best{.904} & \best{.890} & \best{.954} & \best{.030}
& \second{.913} & \second{.874} & \best{.960} & \best{.014}
& \second{.932} & \best{.914} & - & \best{.015}
& \best{.914} & \best{.894} & \best{.953} & \best{.023}\\
SAM2-UNet$_{26}$ \cite{xiong2026sam2_unet} & 352  
& .884 & .861 & .932 & .042
& .880 & .789 & .936 & .021
& .914 & .863 & .961 & .022
& .901 & .863 & .941 & .029\\
\midrule
\rowcolor{samgray}
BED-SAM2 & 352 
& .889 & .852 & .931 & .040
& .885 & .806 & .937 & .020
& .926 & .882 & .960 & .019
& .902 & .858 & .940 & .029\\
\rowcolor{samgray}
BED-SAM2 & 512   
& \second{.898} & .864 & .935 & \second{.038}
& .906 & .845 & .949 & .017
& .928 & \second{.887} & .964 & .019
& .909 & .871 & .942 & \second{.028}\\
\rowcolor{samgray}
BED-SAM2 & 1024 
& \second{.898} & \second{.868} & .934 & .039
& \best{.922} & \best{.875} & \second{.956} & \second{.016}
& \best{.941} & \best{.914} & \second{.969} & \best{.015}
& \second{.912} & \second{.878} & .941 & \second{.028}\\
\bottomrule
\end{tabular}%
} 
\end{table*}

\subsection{RGB SOD}
In Table \ref{tab:SOD_Results}, BED-SAM2 sees improvements across most metrics for each dataset, with some BED-SAM2 model variation achieving the highest S-measure in each of the eight datasets. Some datasets, namely DUTS-TE~\cite{wang2017_DUTS}, benefit from training with all three SOD training datasets, DUTS-TR, HRSOD-TR, and UHRSD-TR. Various image input resolutions make a difference across each dataset; higher resolutions preserve small structures and boundaries that would otherwise be blurred or lost during downsampling. From Table \ref{tab:SOD_Results} we determine that a smaller input resolution of $352 \times 352$ is sufficient for strong results on all datasets, but some datasets perform better at higher input resolutions; namely DUTS-TE, HKU-IS, and PASCAL-S benefit from a larger input resolution of $512 \times 512$ while an input resolution of $1024 \times 1024$ can provide some minor improvements in $F_\beta, E_\phi$ and $\mathcal{M}$, but generally might not be worth the compute cost over more lightweight models.

\subsection{RGB-D SOD}
We have four datasets that provide depth information, as shown in Table \ref{tab:RGBD_Results}. Among these four datasets, we see the best results in the NLPR~\cite{NLPR}, SIP~\cite{Fan_2021_SIP}, and STERE~\cite{6247708_STERE} benchmarks. The reduced performance in NJU2K is due to the dataset containing 3D computer-generated imagery that Distill any Depth struggles to accurately predict. The input resolution has a significant impact on performance in both the high-resolution COD10K~\cite{Fan_2020_CVPR_COD10K} and CHAMELEON~\cite{skurowski2018animal_CHAMELEON} datasets. The most significant jumps in performance are a $3\%$ S-measure improvement in the SIP benchmark and a $1.2\%$ S-measure improvement in STERE.

\subsection{Camouflaged Object Detection}
Our COD results can be found in Table \ref{tab:COD_Results}. While we see improvements in COD10K-TE~\cite{Fan_2020_CVPR_COD10K} and CHAMELEON~\cite{skurowski2018animal_CHAMELEON}, these improvements do not extend to CAMO~\cite{ltnghia-CVIU2019_CAMO} or NC4K~\cite{yunqiu_cod21_nc4k}. Experimental analysis reveals that our edge structure maps are particularly sensitive to noise in heavily textured natural environments, primarily underwater where caustics, high turbidity, and light degradation are common.

Another limitation is evident in the CAMO dataset, where "painted-in" objects are present. Artificially camouflaged subjects (e.g., body painting) effectively deceive monocular depth estimation networks, leading to spurious depth discontinuities in depth maps where none exist.



\section{Conclusion and Future Directions}

In this paper, we propose BED-SAM2, an architecture that demonstrates the capabilities of SAM2 to generalize to unseen modalities. Monocular depth provides complementary information to RGB images, and by taking the edges of that depth, BED-SAM2 can recognize and make confident predictions within those boundaries. We also demonstrate that modern monocular depth estimation can exceed existing ground truth depth data within some RGB-D datasets (e.g., SIP and STERE).

BED-SAM2 achieves state-of-the-art performance in multiple benchmark datasets across three tasks and thirteen datasets. These results highlight the effectiveness of incorporating depth-based edge information in segmentation models.

This work primarily focuses on network improvements through data input with slight architectural modifications. Future work could explore alternative decoder designs, dual-encoder architectures, and mid- or late-fusion strategies to enhance performance further.

\section{Acknowledgments}
Research was sponsored by the Army Research Laboratory and was accomplished under Cooperative Agreement Number W911NF-24-2-0049. The views and conclusions contained in this document are those of the authors and should not be interpreted as representing the official policies, either expressed or implied, of the Army Research Laboratory or the U.S. Government. The U.S. Government is authorized to reproduce and distribute reprints for Government purposes notwithstanding any copyright notation herein.
{
    \small
    \bibliographystyle{ieeenat_fullname}
    \bibliography{main}
}

\end{document}